\documentclass{article}

\usepackage{PRIMEarxiv}

\usepackage[utf8]{inputenc} % allow utf-8 input
\usepackage[T1]{fontenc}    % use 8-bit T1 fonts
\usepackage{hyperref}       % hyperlinks
\usepackage{url}            % simple URL typesetting
\usepackage{booktabs}       % professional-quality tables
\usepackage{amsfonts}       % blackboard math symbols
\usepackage{nicefrac}       % compact symbols for 1/2, etc.
\usepackage{microtype}      % microtypography
\usepackage{lipsum}
\usepackage{fancyhdr}       % header
\usepackage{graphicx}       % graphics
\usepackage{multirow}
\usepackage{amssymb}

\usepackage{amsmath,amsfonts}
\usepackage{mathtools} % automatically loads amsmath
\mathtoolsset{showonlyrefs=true} % only referenced equations are numbered

\usepackage{multirow}

\usepackage{stfloats}
\usepackage{xcolor}
\usepackage{algorithm}  
\usepackage{algorithmic}
\usepackage[normalem]{ulem}
\usepackage{lscape}
\usepackage{longtable}
\usepackage{setspace} 
\usepackage{subfig}
\usepackage{xcolor}
\usepackage{siunitx}

\newcommand{\matrixdim}[2]{\underset{\scriptscriptstyle#2}{#1}}
\DeclareMathOperator{\cat}{cat}
\DeclareMathOperator{\sigmoid}{sigmoid}

\DeclareMathOperator{\cosineSimilarity}{s_{\mathrm{cs}}}

\DeclareMathOperator{\CosineSimilarity}{s_{\mathrm{cs}}}
\newcommand{\T}{\mathrm{T}}

\usepackage[textwidth=1cm, textsize = tiny]{todonotes}

\usepackage[justification=centering]{caption}
 \usepackage{lineno}

\graphicspath{{media/}}     % organize your images and other figures under media/ folder

\title{VITR: Augmenting Vision Transformers with Relation-Focused Learning for Cross-Modal Information Retrieval
}

   \author{
  Yan Gong, Georgina Cosma, Axel Finke \\
  Loughborough University \\
  Loughborough\\
  \texttt{\{y.gong2, g.cosma, a.finke\}@lboro.ac.uk}
  }
  
\begin{document}
\maketitle

\begin{abstract}
  The relations expressed in user queries are vital for cross-modal information retrieval. Relation-focused cross-modal retrieval aims to retrieve information that corresponds to these relations, enabling effective retrieval across different modalities. Pre-trained networks, such as Contrastive Language-Image Pre-training (CLIP), have gained significant attention and acclaim for their exceptional performance in various cross-modal learning tasks. However, the Vision Transformer (ViT) used in these networks is limited in its ability to focus on image region relations. Specifically, ViT is trained to match images with relevant descriptions at the global level, without considering the alignment between image regions and descriptions. This paper introduces VITR, a novel network that enhances ViT by extracting and reasoning about image region relations based on a local encoder. VITR is comprised of two key components. Firstly, it extends the capabilities of ViT-based cross-modal networks by enabling them to extract and reason with region relations present in images. Secondly, VITR incorporates a fusion module that combines the reasoned results with global knowledge to predict similarity scores between images and descriptions. The proposed VITR network was evaluated through experiments on the tasks of relation-focused cross-modal information retrieval. The results derived from the analysis of the RefCOCOg, CLEVR, and Flickr30K datasets demonstrated that the proposed VITR network consistently outperforms state-of-the-art networks in image-to-text and text-to-image retrieval.

\end{abstract}

% keywords can be removed

\section{Introduction}\label{sec:intro}
\begin{figure}[ht]
\centering
\includegraphics[width=0.45\linewidth]{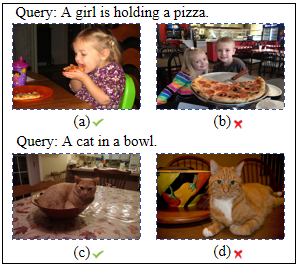}
\caption{A retrieval system that considers region relations will rank the image (a) of `a person holding food' as more relevant to the query description than the image (b) of just `people and food'.}
\label{RelationRetrieval}
\end{figure}

Due to the escalation of multi-modal multimedia data \cite{wei2016heterogeneous, zhang2021probability}, relation-focused cross-modal information retrieval, concentrating on the extraction of information in alignment with relations expressed in user queries, is of particular prominence in the domain of information retrieval applications and the evolution of search engines of the next generation. 
Such capability will result in improved retrieval and ranking performance since the results will be more relevant to the user's query than when relations are not considered. Consider, for example, Figure~\ref{RelationRetrieval}, which shows a description query containing relations, such as `person holding food'. A system that considers relations of image regions will rank images (e.g., Figure~\ref{RelationRetrieval}a) featuring a person holding food as more similar to the query than images (e.g., Figure~\ref{RelationRetrieval}b) depicting people and food separately.

Current works use Visual-Semantic Embedding (VSE) networks to embed image--description pairs in a shared latent space and calculate similarity scores for retrieval tasks \cite{gong2021limitations}. Pre-trained VSE networks have recently gained popularity in various cross-modal tasks \cite{chen2020uniter, yu2021ernie, cheng2022vista, lu2022cots}, with the Contrastive Language-Image Pre-training network (CLIP) \cite{radford2021learning} achieving state-of-the-art performance in cross-modal information retrieval. CLIP employs a pre-trained Vision Transformer (ViT) and a transformer-based text encoder to encode images and descriptions into a shared embedding space. ViTs use the self-attention mechanism from transformers, allowing the model to capture long-range dependencies and intricate patterns in the input data, resulting in a rich contextual understanding of the visual modality and improved cross-modal understanding \cite{mao2021dual}.

ViTs have been extensively studied for cross-modal information retrieval, but there is still room for improvement, particularly in relation-focused tasks. ViTs divide images into small blocks \cite{liu2021dimbert}, which can result in a loss of local information compared to Convolutional Neural Networks (CNNs) \cite{yan2016cnn}. This limitation becomes apparent when applying ViT-based pre-trained VSE networks to relation-focused tasks, as the models exhibit weak local perception abilities for images and have limited capacity to align image regions with corresponding descriptions. Additionally, the ViT used in contrastive learning \cite{radford2021learning} connects with a convolutional layer, its primary design objective is to capture global image features rather than aligning image regions with corresponding descriptions \cite{mao2021dual, gal2022stylegan}. Modifying the internal structure of the transformer to enhance ViTs ability for capturing local image information may result in potential drawbacks in cross-modal tasks, such as increased computational complexity, a larger number of parameters, and the risk of losing global context. 

To address the limitations of ViTs in relation-focused cross-modal information retrieval tasks, this paper proposes a novel network named VITR. 
VITR provides relational reasoning of image regions that are extracted by a local encoder, and fuses these relations into the pre-trained Vision Transformer for relation-focused cross-modal information retrieval tasks. In this paper, relational reasoning involves extracting relevant relations between image regions and generating relation-focused local representations of the image to improve cross-modal information retrieval performance. The contributions of this paper are as follows:
\begin{itemize}
    \item The proposed VITR network benefits from a novel fusion module that fuses the global and local representations of images and descriptions to predict the similarity scores of image and description pairs. VITR utilises a ViT encoder and a text encoder to derive the global representations of image and description pairs, and a CNN-based local encoder to determine the local representations of image and description pairs.
    
    \item VITR leverages a new type of relational reasoning module that first models an image's regions and their relations using a relational graph, then generates local representations aligned with the image's description. Incorporating relation-focused local image representations into VITR improves cross-modal information retrieval performance.

    \item This paper also enhances the information retrieval efficiency (time) of VITR, an aspect often overlooked by current cross-modal networks. It proposes a module, called turbo, for VITR, which selects the top $N$ relevant candidates to the query and sends the necessary candidate embeddings or global representations to relevant modules for further finalisation of ranking. The turbo results in reduced computation time of VITR.
    
    \item Extensive experiments were carried out by evaluating VITR to the datasets RefCOCOg and CLEVR which involve relation-focused descriptions, and the benchmark dataset Flickr30K. VITR outperformed various other state-of-the-art networks,  namely CLIP, VSRN++, and VSE$\infty$, in both image-to-text and text-to-image cross-modal information retrieval tasks.
\end{itemize}

The rest of the paper is organised as follows: Section~\ref{sec:relawork} summarises the related work, Section~\ref{sec:methodology} elaborates the proposed VITR, Section~\ref{sec:experimets} demonstrates the experimental results, Section~\ref{sec:visualisation} presents the visualisation results, and Section~\ref{sec:conclusion} concludes our work.

\section{Related Work}
\label{sec:relawork}
\subsection{Visual Semantic Embeddings}
% Wang~et~al.~\cite{wang2020pfan++} introduced a novel position-focused attention network for bi-directional image-text retrieval and matching, enhancing visual-text joint-embedding learning by integrating object position information. 
Faghri et al. \cite{faghri2018vse++} unveiled VSE++, an elevated Visual-Semantic Embedding architecture that incorporates a fully connected neural network to generate the representations of image features extracted by a faster R-CNN \cite{ren2016faster} and a GRU network \cite{cho2014learning} to generate the representations of descriptions. 
Wang~et~al.~\cite{wang2023rare} introduced a rare-aware attention network, which aims to address the long-tail effect in image and text matching by exploring and exploiting rare textual content. 
Lee et al. \cite{lee2018stacked} introduced an attention network designed to unveil the complete latent alignments between image regions and their respective descriptive words. 
Li et al.\ \cite{li2019visual} introduced the Visual Semantic Reasoning Network (VSRN), designed to augment image features using image region relationships, these relationships being extracted via a Graph Convolutional Network (GCN) \cite{zhang2019graph}.  
Later, Li et al.\ \cite{li2022image} improved the VSRN by upgrading it to VSRN++, which replaces the word2vector embeddings with pre-trained BERT \cite{devlin2019bert} embeddings.
Chen et al.\ \cite{chen2021learning} proposed a variant of the VSE network, VSE$\infty$, which leverages a generalised pooling operator to discern the most effective strategy for pooling the representations of images and descriptions.

\subsection{Pre-trained Networks for Visual Semantic Embeddings}
The development of pre-trained networks for cross-modal information retrieval has progressed significantly in recent years \cite{chen2020uniter, yu2021ernie, cheng2022vista, gan2020large, lu2022cots, zhang2023universal}. 
Chen et al. \cite{chen2020uniter} presented UNITER, a model that serves as a universal image-text bridge, meticulously pre-trained on four distinct image-text datasets. This network accommodates a diverse array of vision-and-language tasks, generating joint multimodal embeddings through four dedicated pre-training tasks.  
Yu et al. \cite{yu2021ernie} put forth a methodology that leverages structured knowledge from scene graphs to boost joint representation learning in tasks that intersect vision and language. 
Lu et al.\ \cite{lu2022cots} proposed a novel collaborative two-stream vision-language pre-training approach for image-text retrieval that enhances cross-modal interaction through instance-level alignment, token-level interaction, and task-level interaction.
Recently, Radford et al.\ \cite{radford2021learning} proposed the pre-trained CLIP which applies contrastive learning to align the global visual representations and textual representations from a dataset including 400 million image--description pairs. 
The architecture of CLIP involves: (1) a text encoder which aims to embed the description as a dimension-reduced representation; (2) an image encoder, commonly using ViT, which aims to embed the image as a representation with the same dimension as the description representation. 
% Radford et al.\ \cite{radford2021learning} also provided a CNN version of the image encoder. However, due to the advantage of ViT on pre-training, the ViT encoder outperformed the CNN encoder, and it is most commonly applied in other works \cite{ma2022ei}. 
CLIP has been applied in many tasks recently, such as e-commerce image retrieval \cite{ma2022ei}, video-text retrieval \cite{ma2022x}, and text-image generation \cite{tao2023galip}. However, the pre-trained networks, especially CLIP, still lack the ability to effectively match local information in images to their descriptions in cross-modal information retrieval tasks.

\subsection{Relational Reasoning Methods}
Graphs are invaluable for representing and analysing relations \cite{lee2019attention, ke2022multi, maurya2021graph}. 
In recent years, graph-based methods have shown an efficient way of reasoning with relations \cite{yoon2021image, zhang2023boosting, wang2021dualvgr, wang2022multi, chen2023multi}.
% Yu et al.\ \cite{yu2020cross} proposed a network based on both the visual and semantic graphs for visual question answering.
For a scene graph generation task, Lin et al.\ \cite{lin2022atom} explored the atom correlation-based graph propagation which incorporates prior knowledge in a more stable and comprehensive way;
and Cuiet al.\ \cite{cui2018context} propose a framework for visual relationship detection that uses word semantic and visual scene graphs to capture global context interdependency among object instances. 
For cross-modal information retrieval, Cao et al.\ \cite{cao2021global} introduced a graph-based relation-aware attention module to weigh image fragments based on the pairwise relations of the fragments; 
and Li et al.\ \cite{li2022image} applied a GCN to extract relations between image regions, and used the extracted relations to enhance image features. 

\section{Proposed ViT-Relation-Focus (VITR) Network}
\label{sec:methodology}
\begin{figure*}[htbp]
\centering
\includegraphics[width=0.98\linewidth]{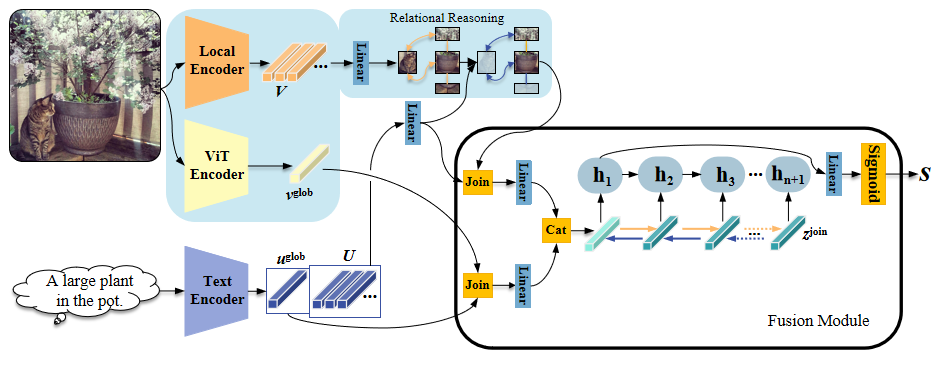}
\caption{An overview of the proposed VITR. VITR consists of: (1) a pre-trained \emph{text encoder} that provide pre-trained language knowledge of an image's description; (2) a pre-trained \emph{ViT encoder} that encodes an image as a global representation, and a CNN-based \emph{local encoder} that extracts features from image regions; (3) a \emph{relational reasoning} module that models the relations between regions in an image and generates local representations of the regions based on their descriptions; and (4) a \emph{fusion} module that fuses the outputs from relational reasoning and pre-trained knowledge through a sequence-optimised graph network, and outputs the similarity score between the image $I$ and description $D$.}
\label{VITR}
\end{figure*}

\textbf{Overview.}
The proposed VITR network is illustrated in Figure~\ref{VITR}. Given an image $I$ and a description $D$, VITR aims to embed the pair $(I, D)$ into the shared latent space for predicting its similarity score $s(I,D)$. 
VITR is comprised of:
(1) A text encoder which encodes the description $D$ to incorporate pre-trained language knowledge. 
(2) A ViT encoder and a CNN-based local encoder which encode the image $I$ and its regions as a global representation and a set of features respectively. 
(3) A relational reasoning module that represents image regions in relations, and generates local representations of the image regions based on their descriptions.
(4) A fusion module that predicts the similarity score $s(I,D)$ based on fusing the results of VITR's relational reasoning module and pre-trained knowledge using a sequence-optimised graph network.

\subsection{Encoding the Description}
VITR utilises a pre-trained text encoder (e.g.\ CLIP's text encoder \cite{radford2021learning} or pre-trained BERT \cite{chen2021learning}). This module encodes the description $D$ as a global representation vector $u^{\mathrm{glob}}\in \mathbb{R}^{d_{1}}$, and a collection of word embedding vectors  $U = \{u_1, \dotsc, u_n\}$, where $n$ is the number of words in the description, and $u_{j} \in \mathbb{R}^{d_1}$ is the $j$th word embedding vector with dimension $d_{1}$.
The output of this module is $(U, u^{\mathrm{glob}})$.

\subsection{Encoding the Image} \label{Detection Module}
VITR encodes the image using two components. The ViT encoder utilises a pre-trained vision transformer network based on cross-modal learning, such as the image encoder of CLIP's ViT model. This module encodes an image $I$ as a global representation vector $v^{\mathrm{glob}}\in \mathbb{R}^{d_{1}}$. The local encoder utilises a pre-trained CNN to encode the image $I$ into a set of regional representations:
\begin{align}
    \mathrm{CNN}(I) = V = \{v_1, \dotsc, v_k\},
\end{align}
where each feature $v_{i} \in \mathbb{R}^{d_2}$ encodes a salient region of the image and $k$ is the total number of regions. The output of this module is $(V, v^{\mathrm{glob}})$. Examples of such CNN networks include the image encoder of CLIP's ResNet model \cite{radford2021learning} or the ResNet backbone of Faster-RCNN \cite{ren2016faster}.

\subsection{Proposed Relational Reasoning} \label{Reasoning Module}
% \textbf{Preparation for Unifying the Dimensions of Vectors.}
For further computation with crossing modalities, the elements of $V$ and $U$ are projected into a unified dimension $d_{3}$ as follows:
\begin{align}
 v_i^* & = W^{v}(v_i),\\
 u_j^* & = W^{u}(u_j),
\label{equationPVU}
\end{align}
where the weight parameters $W^{v}$ and $W^{u}$ are both the fully connected layers with $d_{3}$ output neurons. Here, $v^{*}_{i} \in \mathbb{R}^{d_3}$ is the projected vector corresponding to the $i$th region. Similarly, $u^{*}_{j} \in \mathbb{R}^{d_3}$ is the $j$th embedding vector for the $j$th word in the description. Finally, set $V^* = \{v_1^*, \dotsc, v_k^*\}$ and $U^* = \{u_1^*, \dotsc, u_n^*\}$.

% \textbf{Modeling Relations.} 
% This block aims to represent an image's regions and their relations extracted by a multi-layer graph neural network.
The regions of an image and their relationships are represented using a multi-layer graph neural network. Let $R \in \mathbb{R}^{k \times k}$ be a matrix of relations of regions and computed whose element $(i, l)$, for any $1 \leq i, l \leq k$, is:
\begin{equation}
  [R]_{i,l} = W^{\varphi_{1}}(v_i^*)^\T W^{\phi_{1}}(v_l^*),
\label{equationR}
\end{equation}
where the weight parameters $W^{\varphi_{1}}$ and $W^{\phi_{1}}$ are both fully connected layers with $d_{3}$ output neurons. From this, a matrix of pairwise relations of regions is computed as:
\begin{equation}
  \matrixdim{R^{\mathrlap{\mathrm{pw}}}}{k \times k} \;\;= \sigma (W^{\mathrm{R}_1}(\cat[R^{\mathrm{T}}, R])), 
\label{equationRp}
\end{equation}
where $\cat$ denotes row-wise concatenation and $\sigma$ denotes the $\tanh$ activation function. The weight parameter $W^{\mathrm{R}_1}$ is a 1D convolutional layer (kernel size $1$; $k$ output channels). Additionally, let $R^{\mathrm{inn}} = (r_1^{\mathrm{inn}}, \dotsc, r_k^{\mathrm{inn}}) \in \mathbb{R}^{k}$ hold the inner information for each vector in $V^{*}$:
\begin{equation}
  r_i^{\mathrm{inn}} = \sigma (W^{\mathrm{R}_2}(v_i^*)) \in \mathbb{R}^{k},
\label{equationRs}
\end{equation}
where the weight parameter $W^{\mathrm{R}_2}$ is a fully connected layer with one output neuron. Merge $R^{\mathrm{pw}}$ and $R^{\mathrm{inn}}$ into $R^{\mathrm{a}}$ as:
\begin{equation}
  \matrixdim{R^{\mathrm{a}}}{k \times 1} = \sigma (W^{\mathrm{R}_3}(\cat[R^{\mathrm{pw}}, R^{\mathrm{inn}}])),
\label{equationRf}
\end{equation}
where $\cat$ now denotes column-wise concatenation; the weight parameter $W^{\mathrm{R}_3}$ is a fully connected layer with one output neuron. 
A collection of representation vectors $V^{\mathrm{a}} = \{v_1^{\mathrm{a}}, \dotsc, v_k^{\mathrm{a}}\}$ for the $k$ regions is then obtained as
\begin{equation}
 v_i^{\mathrm{a}} = \sigmoid([R^{\mathrm{a}}]_i)v_i^* \in \mathbb{R}^{d_3},
\label{equationV'}
\end{equation}
where $[R^{\mathrm{a}}]_i$ is $i$th element of $R^{\mathrm{a}}$.
Finally, Equations~\refeq{equationR}--\refeq{equationV'} can be recursively repeated $g_{1}\in \mathbb{N}^{+}$ times ($g_1 = 4$ in this paper). In this case, the output $V^{\mathrm{a}}$ from the repetition forms the input $V^*$.

% \textbf{Relations Expressed in Text.}
Since not all visual vectors in the set $V^{\mathrm{a}}$ are relevant to the description, the visual vectors are weighted to generate local representations of the image that are aligned with the descriptive words, denoted $V^{\mathrm{rela}} = \{v_1^{\mathrm{rela}}, \dotsc, v_n^{\mathrm{rela}}\}$. Here, the newly generated image local representation vector aligned with the $j$th word is given by
\begin{equation}
v^{\mathrm{rela}}_{j} = \sum_{i=1}^{k}a_{i,j} v_i^{\mathrm{a}},
\label{equationvstar}
\end{equation}
where
\begin{align}
  a_{i,j} = \frac{\exp(\gamma  \bar{s}_{i,j})}{\sum_{i'=1}^k \exp(\gamma  \bar{s}_{i',j})},\label{equationaij}
\end{align}
are weights that are specified through a softmax function with inverse temperature parameter $\gamma > 0$ (set to $\gamma = 12$ by this paper). Here, 
\begin{align}
  \bar{s}_{i,j} = \frac{[\CosineSimilarity(v_{i}^{\mathrm{a}}, u^{*}_{j})]_{+}}{\sqrt{\sum_{j'=1}^{n}[\CosineSimilarity(v_{i}^{\mathrm{a}}, u^{*}_{j'})]_{+}^{2}}},
\end{align}
with $[x]_{+} = \textrm{max}(x,0)$, is a normalised and thresholded version of the cosine similarity $\cosineSimilarity$. In summary, the output of the relational reasoning module is $(V^{\mathrm{rela}}, U^{*})$.

\subsection{Proposed Fusion Module} \label{Fusion module}
This module predicts the similarity score $s(I,D)$ for the image--description pair by fusing the results of the relational reasoning module and the global representations of the image and description. The process is described as follows.

% \textbf{Joint Image--Description Representations.}
% This operation aims to combine the local and global image--description representations and embed them in the same low-dimensional (intended to reduce computational complexity) latent space for fusion processing.
The local and global image-description representations are combined and embedded in the same low-dimensional latent space (intended to reduce computational complexity) for fusion processing.  
More formally, a vector $z^{\mathrm{glob}}$ for joining a global image--description representation pair $(v^{\mathrm{glob}}, u^{\mathrm{glob}})$, and vectors $z^{\mathrm{rela}}_{j}$ for joining local image--description representation pairs $(v^{\mathrm{rela}}_{j}, u^{*}_{j})$ are computed as:
\begin{align}
    z^{\mathrm{glob}} & = (u^{\mathrm{glob}}-v^{\mathrm{glob}})^{2},
    \\z^{\mathrm{rela}}_{j} & = (u^{*}_{j}-v^{\mathrm{rela}}_{j})^{2},
    \label{equationssclip}
\end{align}
where $(\, \cdot \, )^{2}$ is applied element-wise. Furthermore, define the vectors (in $\mathbb{R}^{d_4}$):

\begin{align}
  z_0^{\mathrm{join}} &= W^{\mathrm{glob}}(z^{\mathrm{glob}}),\\
  z_i^{\mathrm{join}} &= W^{\mathrm{rela}}(z_i^{\mathrm{rela}}), \quad \text{for $i = 1, \dotsc, n$,}
\end{align}
where the weight parameters $W^{\mathrm{rela}}$ and $W^{\mathrm{glob}}$ are both fully connected layers with $d_{4}$ (e.g.\ 128) output neurons. 

% \textbf{Fusing Contextual Information to the Combined Representation.}
To ensure that $z^{\mathrm{join}}_{j}$ contains sufficient contextual information, it can be treated as a node for constructing a graph. The edge matrix $E = 
%(e_{j,l}) \in 
\mathbb{R}^{(n+1) \times (n+1)}$ is obtained (for any $1 \leq j, l \leq (n+1)$) as:
\begin{equation}
  % e_{j,l} 
  [E]_{j,l} = W^{\varphi_{2}}(z_j^{\mathrm{join}})^\T W^{\phi_{2}}(z_l^{\mathrm{join}}),
\label{equationsE}
\end{equation}
where the weight parameters $W^{\varphi_{2}}$ and $W^{\phi_{2}}$ are both fully connected layers with $d_{4}$ output neurons. Then the information among the joined vectors is fused as:
\begin{equation}
z_j^{\mathrm{fuse}} = \sum_{l=1}^{n+1} W^{\mathrm{fuse}}(\sigmoid([E]_{j,l}) z_j^{\mathrm{join}}),
\label{equationsSfuse}
\end{equation}
where the weight parameter $W^{\mathrm{fuse}}$ is a fully connected layer with $d_{4}$ output neurons. Finally, set $Z^{\mathrm{fuse}} = \{z_1^{\mathrm{fuse}}, \dotsc, z_{n+1}^{\mathrm{fuse}}\}$. Equations~\refeq{equationsE}--\refeq{equationsSfuse} can be recursively repeated $g_{2}\in \mathbb{N}^{+}$ times ($g_2 = 2$ in this paper), where the output $Z^{\mathrm{fuse}}$ from the last time is taken as the input for the next time. A sequence optimiser is utilised to dynamically capture and incorporate the temporal dependencies among the elements of $Z^{\mathrm{fuse}}$. This allows the module to generate a rich and complex combined representation of $Z^{\mathrm{fuse}}$ as: 
\begin{equation}
\{h_{j}\}_{j=1}^{n+1} =\textrm{GRU}(Z^{\mathrm{fuse}}),
\label{equationsGRU}
\end{equation}
where $\{h_{j}\}_{j=1}^{n+1}$ are the hidden states of a GRU layer, and only $h_{1}$ is taken as the combined representation of $Z^{\mathrm{fuse}}$.

% \textbf{Predicting Image--Description Similarity Score.}
Finally, the similarity score $s$ for a pair $(I,D)$ is predicted as:
\begin{equation}
s(I, D) = \sigmoid(W^{\mathrm{h}}(h_{1})),
\label{equationssfinal}
\end{equation}
where the weight parameter $W^{\mathrm{h}}$ is a fully connected layer with one output neuron. %, and the result $s(I,D) \in[0,1]$ is predicted by the sigmoid function with the input $h_{1}$ weighted by $W^{\mathrm{h}}$.

\subsection{Training VITR}
The pre-trained models - text, ViT, and local encoders - constitute an integral part of the VITR framework. The remaining parameters within VITR undergo a collaborative training process facilitated by LSEH \cite{gong2023improving}. LSEH, which serves as an advanced version of the hard negatives loss function, focuses on learning the distances between image-description pairs \cite{faghri2018vse++}. 
Consider $\{(I_{1}, D_{1}), \dotsc, (I_{m}, D_{m})\}$ as a training dataset consisting of image--description pairs. Each image $I_{p}$ is associated with its corresponding relevant description $D_{p}$, where $p$ denotes the pair index, and $m$ represents the total number of pairs in the training set.  
Given a relevant image--description pair $(I_{p},D_{p})$, the result of LSEH only takes the max from the irrelevant pairs as:
\begin{alignat}{2}
   \mathrm{L}(I_{p},D_{p}) = & \max_{\hat{D}_{p}}[\alpha+\lambda \cosineSimilarity(D^{'}_{p}, \hat{D}^{'}_p)+s(I_{p},\hat{D}_{p})-s(I_{p},D_{p})]_{+} +\\
   & \max_{\hat{I}_p}[\alpha+\lambda \cosineSimilarity(D^{'}_{p}, \hat{D}^{'}_p)+s(D_{p},\hat{I}_p)-s(I_{p},D_{p})]_{+},
   \label{equationLSEH1}
\end{alignat}

where $\alpha$ (set to 0.185) is a margin parameter and $\lambda$ (set to 0.025) is a temperature parameter. Furthermore, $\hat{D_{p}}$ and $\hat{I}_p$ are irrelevant images and descriptions, respectively (e.g.\ from the entire data set or a mini batch). Additionally, the semantic factors $\smash{\lambda \cosineSimilarity(D^{'}_{p}, \hat{D}^{'}_p)}$ dynamically adjust the margin $\alpha$ according to the cosine similarity between $D^{'}_{p}$ and $\hat{D}^{'}_p$ for flexible learning of the network. 

The terms $D^{'}_{p}$ and $\hat{D}^{'}_{p}$ are decomposition eigenvalues of $D_{p}$ and $\hat{D}_{p}$ respectively which are obtained as follows. Define the matrix $A = \cat[D_1^\T, \dotsc, D_m^\T] \in \mathbb{R}^{m \times w}$, where $m$ is the number of descriptions, $w$ denotes the total count of unique terms present in the description set, and $\cat$ denotes row-wise concatenation. Then truncated SVD is applied to $A$ as follows:
\begin{align}
    \matrixdim{A}{m \times w} \approx \matrixdim{X}{m \times d_{5}} \; \matrixdim{\Lambda}{d_{5}\times d_{5}} \; \matrixdim{Y^\T}{d_{5}\times w},\quad 
    \matrixdim{B}{m\times d_{5}} = \matrixdim{A}{m\times w} \; \matrixdim{Y}{w\times d_{5}},
    \label{equationSVD}
\end{align}
where $d_{5}$ is the number of singular values. The $m$ rows of the matrix $B$ are the vectors $D^{'}_{1}, \dotsc, D^{'}_{m} \in \mathbb{R}^{d_5}$, of reduced representations of descriptions. The reduced representations of irrelevant descriptions, $\hat{D}'_p$, are then taken from the set $\{D^{'}_{1}, \dotsc, D^{'}_{m}\}$.

\subsection{Proposed Turbo Module for Improving Retrieval Efficiency} \label{turbo}

\begin{figure}[ht]
\centering
\includegraphics[width=0.5\linewidth]{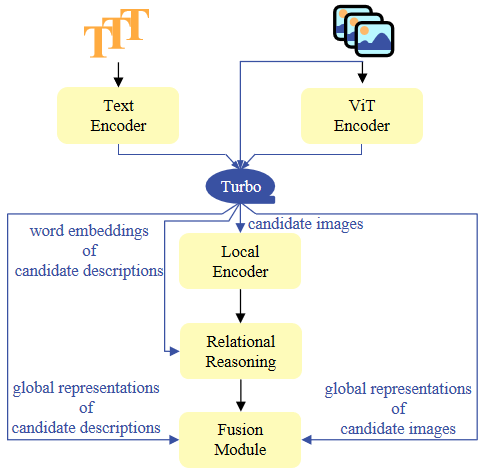}
\caption{The proposed \emph{turbo} module for VITR selects the top $N$ candidate descriptions or images for the query, and sends the necessary candidates' embeddings or representations to the relevant modules for further finalisation of ranking.}
\label{VITR_turbo}
\end{figure}

Aimed at improving information retrieval efficiency (time), this paper proposes a turbo module for VITR that can be used exclusively for retrieval (not for training). The proposed turbo module is shown in Figure~\ref{VITR_turbo}: 

\textbf{Input.} The turbo receives the images, the output of the ViT encoder (i.e., the global representation of the images), and the output of the text encoder (i.e., the global representations of the descriptions).

\textbf{Operation.} The turbo ranks the descriptions based on the cosine similarities between the query image's global representation and the global representations of the descriptions for image-to-text retrieval; or ranks the images based on the cosine similarities between the query description's global representation with the images' global representations for text-to-image retrieval. It then selects the top $N$ ($N\in \mathbb{N}^{+}$) relevant candidates for the query based on the ranking results.

\textbf{Output.} For image-to-text retrieval, turbo sends the candidate descriptions' word embeddings and global representations to the relational reasoning and fusion modules, respectively, and the query image and its global representation to the local encoder and the fusion modules, respectively.
For text-to-image retrieval, turbo sends the candidate images and their global representations to the local encoder and the fusion module, respectively, and the query description's word embeddings and global representation to the relational reasoning and fusion modules, respectively.

Finally, each module in VITR performs computations based on the received results from turbo to finalise the ranking of candidate descriptions or images for the query. By using the turbo module, the computational complexity of the major components of VITR (involving the local encoder, and the relational reasoning and fusion modules) is reduced by a factor of $\frac{length}{N}$, where $length$ is the length of the database, and $N$ is the number of candidates.

\section{Experiments}
\label{sec:experimets}
% The proposed VITR was evaluated in the datasets of Flickr30K \cite{young2014image}, RefCOCOg \cite{mao2016generation}, and CLEVR \cite{johnson2017clevr} for the tasks of image-to-text and text-to-image retrieval, and the performance of VITR is compared to state-of-the-art networks.
The proposed VITR underwent evaluation using the Flickr30K \cite{young2014image}, RefCOCOg \cite{mao2016generation}, and CLEVR \cite{johnson2017clevr} datasets for image-to-text and text-to-image retrieval tasks. The performance of VITR was then benchmarked against that of state-of-the-art networks.

\subsection{Evaluation Measures and Datasets}
The evaluation metric used for the cross-modal information retrieval experiments is Recall at rank $K$ (Recall@$K$), which measures the percentage of relevant items included in the top $K$ retrieved results \cite{gong2021limitations}. The experiments aim to evaluate the network's ability to retrieve at least one relevant item from a given list of relevant items, and the average Recall is computed across the results of the evaluated queries.

% \subsection{Datasets}
The RefCOCOg, CLEVR, and Flickr30K datasets are split as shown in Table~\ref{DatasetsSplit} and described as follows.

\begin{table}[htbp]
\caption{Dataset split of RefCOCOg, CLEVR, and Flickr30K.}
\label{DatasetsSplit}
\centering
% \resizebox{0.7\linewidth}{!}{
\begin{tabular}{lcccc}
\hline
Dataset   & \multicolumn{1}{c}{Modality}                                   & Train                                                  & Validate                                            & Test                                                \\ \hline
RefCOCOg  & \begin{tabular}[c]{@{}c@{}}images\\ descriptions\end{tabular} & \begin{tabular}[c]{@{}c@{}}21899\\ 80512\end{tabular}  & \begin{tabular}[c]{@{}c@{}}1300\\ 4896\end{tabular} & \begin{tabular}[c]{@{}c@{}}2600\\ 9582\end{tabular} \\ 
CLEVR    & \begin{tabular}[c]{@{}c@{}}images\\ descriptions\end{tabular} & \begin{tabular}[c]{@{}c@{}}30000\\ 98345\end{tabular}  & \begin{tabular}[c]{@{}c@{}}1000\\ 3136\end{tabular} & \begin{tabular}[c]{@{}c@{}}1000\\ 3121\end{tabular} \\ 
Flickr30K & \begin{tabular}[c]{@{}c@{}}images\\ descriptions\end{tabular} & \begin{tabular}[c]{@{}c@{}}29000\\ 145000\end{tabular} & \begin{tabular}[c]{@{}c@{}}1000\\ 5000\end{tabular} & \begin{tabular}[c]{@{}c@{}}1000\\ 5000\end{tabular} \\ \hline
\end{tabular}
% }
\end{table}

The \textbf{RefCOCOg} dataset \cite{mao2016generation} contains real-world images taken from the MS-COCO dataset \cite{lin2014microsoft} and their corresponding descriptions provided by the University of Maryland (UMD). The UMD descriptions of RefCOCOg contain information on relations between regions in images, so RefCOCOg is more challenging than MS-COCO for cross-modal information retrieval. On average, each image in the dataset is associated with four relevant descriptions. 

The \textbf{CLEVR} dataset \cite{johnson2017clevr} consists of images depicting 3D-rendered objects. Since this dataset has not been specifically tailored for relation-focused cross-modal retrieval tasks, image descriptions were formulated using the given relational annotations like 'left', 'right', 'front', and 'behind'. 
The dataset was then split into train, test, and validation sets. 
% The prepared image descriptions and the indices of the images found in the train, test, and validation sets 
% are provided in our GitHub repository for experiment reproducibility purposes.\footnote{Once the paper is accepted, the GitHub repository will be made public.\label{github}} 
On average, each CLEVR image is associated with three relevant descriptions. An example description is `A large blue metal cube is behind a large blue rubber sphere'. 

The \textbf{Flickr30K} dataset \cite{young2014image} is a commonly used benchmark for evaluating the performance of VSE networks \cite{radford2021learning, chen2021learning, li2022image}. Each image in the dataset is associated with five textual descriptions.

\subsection{Implementation Details}
All experiments were conducted on a workstation with NVIDIA RTX3090 GPU with PyTorch framework. 
% and the source code files of the experiments are provided in our GitHub repository.\textsuperscript{\ref{github}} 
The networks were implemented as follows.

\textbf{Baselines: CLIP models.\label{Baseline: CLIP.}}
Three CLIP baseline models were selected. These were the base ViT model (`ViT-B/16' with dimension $d_{1}$ of 512), the large ViT model (`ViT-L/14' with dimension $d_{1}$ of 768), and the Resnet101 model (`RN101') \cite{radford2021learning} denoted as $\textrm{CLIP}_{\textrm{B16}}$, $\textrm{CLIP}_{\textrm{L14}}$, and $\textrm{CLIP}_{\textrm{RN101}}$ respectively. Each model was fine-tuned for each dataset to present its best performance, and the hyperparameter settings follow each model's benchmark settings \cite{radford2021learning}.  

\textbf{Experiment Setup of VITR.}
VITR was implemented using the ViT and text encoders from the fine-tuned $\textrm{CLIP}_{\textrm{B16}}$ and $\textrm{CLIP}_{\textrm{L14}}$ models, this resulted in two models of VITR network, namely $\textrm{VITR}_{\textrm{B}}$ and $\textrm{VITR}_{\textrm{L}}$, respectively.
The image encoder of the fine-tuned $\textrm{CLIP}_{\textrm{RN101}}$ was applied for encoding image regions for both $\textrm{VITR}_{\textrm{B}}$ and $\textrm{VITR}_{\textrm{L}}$, and it extracts 49 features (with dimension $d_{2}$ of 2048) of regions from each image. 
Both $\textrm{VITR}{\textrm{B}}$ and $\textrm{VITR}{\textrm{L}}$ underwent training on each dataset for 20 epochs, with a set batch size of 128. The learning rate was fixed at 0.0004 and was subjected to a decay rate of 0.1, commencing at the 5th epoch. This training process made use of the Adam optimizer.

\textbf{Additional Baselines.} 
Two additional baseline networks were chosen for comparison:
(1) VSE$\infty$, which is a representation pooling network \cite{chen2021learning}.
(2) VSRN++, a network that focuses on reasoning image relations \cite{li2022image}. 
Those two networks were implemented and tuned to achieve their best performance on the RefCOCOg and CLEVR datasets. 
For a fair comparison, the two networks used the same extracted features of image regions as VITR. The hyperparameters of each network for the RefCOCOg and CLEVR datasets refer to each network are based on the benchmark settings established for the Flickr30K dataset \cite{chen2021learning, li2022image}.

\subsection{Results}

\textbf{Results on RefCOCOg.}
Table~\ref{RefCOCOgResults} compares the proposed VITR with the baseline methods on the RefCOCOg test set for cross-modal information retrieval, and the main findings are described as follows.
$\textrm{VITR}_{\textrm{L}}$ reached a Recall@1 of 45.2\,\% for image-to-text, and a Recall@1 of 29.5\,\% for text-to-image retrieval. 
Observing the performance of the networks using the Recall@1 metric, $\textrm{VITR}_{\textrm{L}}$ outperformed $\textrm{CLIP}_{\textrm{L14}}$ by 2.8\,\% and 4.3\,\% for image-to-text and text-to-image retrieval respectively, and also outperformed VSE$\infty$ by 14.1\,\% and 10\,\% for those tasks, respectively.
$\textrm{VITR}_{\textrm{B}}$ reached a Recall@1 of 42.9\,\%  and 27.9\,\%  for image-to-text and text-to-image retrieval respectively, and outperformed $\textrm{CLIP}_{\textrm{B16}}$ by 3.6\,\% and 4.1\,\% for those tasks, respectively.

\begin{table}[ht]
\caption{Results of cross-modal information retrieval networks on the RefCOCOg test set. Table shows average Recall@$K$ (\%) values.}
\label{RefCOCOgResults}
\centering
% \resizebox{\linewidth}{!}{
\begin{tabular}{lcccccc}
\hline
\multirow{2}{*}{Network} & \multicolumn{3}{c}{Image-to-Text} & \multicolumn{3}{c}{Text-to-Image} \\
\cmidrule(lr){2-4} \cmidrule(lr){5-7} 
                         & Recall@1           & Recall@5           & Recall@10          & Recall@1           & Recall@5           & Recall@10          \\ \hline
VSRN++                   & 20.0          & 44.9          & 57.3          & 13.8          & 34.6          & 47.8          \\
% SGRAF                   &  26.0             & 53.1              & 65.6               &  17.6             & 40.4              &  52.6           \\
VSE$\infty$              & 31.1          & 58.3            & 69.7          & 19.5          & 42.8         & 55.2           \\
$\textrm{CLIP}_{\textrm{RN101}}$                     & 36.3           & 61.3          & 71.2           & 20.8            & 44.2           & 56.7          \\
$\textrm{CLIP}_{\textrm{B16}}$                     & 39.3            & 64.3           & 75.0          & 23.8            & 48.4            & 60.4          \\ 
$\textrm{CLIP}_{\textrm{L14}}$                     & 42.4            & 65.5           & 75.1         & 25.2           & 48.9           & 60.4           \\
\hline
$\textrm{VITR}_{\textrm{B}}$                 & 42.9  & 68.2   & 79.2 & 27.9  & 53.5  & 65.6 \\
$\textrm{VITR}_{\textrm{L}}$                 & \textbf{45.2} & \textbf{71.1} & \textbf{80.5} & \textbf{29.5} & \textbf{55.1} & \textbf{66.8} \\ \hline
\end{tabular}
% }
\end{table}

\textbf{Results on CLEVR.}
Table~\ref{CLEVRResults} compares VITR with the baseline methods on the CLEVR test set for cross-modal information retrieval, and the main findings are described as follows.
For $\textrm{VITR}_{\textrm{L}}$, Recall@1 reached 90.7\,\% for image-to-text and 79.3\,\% for text-to-image retrieval, and outperformed $\textrm{CLIP}_{\textrm{L14}}$ by 25.1\,\% and 14.1\,\% for those tasks respectively.
$\textrm{VITR}_{\textrm{L}}$'s Recall@1 also outperformed VSE$\infty$'s Recall@1 by 22.9\,\% and 8.5\,\% for image-to-text and text-to-image retrieval respectively. 
The Recall@1 values of $\textrm{VITR}_{\textrm{B}}$ were 88.3\,\%  and 79.4\,\% for image-to-text and text-to-image retrieval respectively. The Recall@1 values of $\textrm{VITR}_{\textrm{B}}$ outperformed that of $\textrm{CLIP}_{\textrm{B16}}$ by 21.5\,\% for image-to-text and 13.9\,\% for text-to-image retrieval respectively.

\begin{table}[ht]
\caption{Results of cross-modal information retrieval networks on the CLEVR test set. Table shows average Recall@$K$ (\%) values.}
\label{CLEVRResults}
% \resizebox{\linewidth}{!}{
\centering
\begin{tabular}{l S[table-format=2.1] S[table-format=2.1] S[table-format=3.1]  S[table-format=2.1] S[table-format=2.1] S[table-format=2.1]}
% \begin{tabular}{lcccccc}
\hline
\multirow{2}{*}{Network} & \multicolumn{3}{c}{Image-to-Text} & \multicolumn{3}{c}{Text-to-Image} \\
\cmidrule(lr){2-4} \cmidrule(lr){5-7} 
                         & {Recall@1} & {Recall@5} & {Recall@10}       & {Recall@1} & {Recall@5} & {Recall@10}       \\ \hline
VSRN++                   & 64.3               & 96.9             & 99.4              & 58.5              & 91.0              & 96.1               \\
$\textrm{CLIP}_{\textrm{RN101}}$                     & 65.4          & 98.1             & 99.8          &  61.8            & 95.7           & 98.1           \\
$\textrm{CLIP}_{\textrm{B16}}$                     & 66.8          & 98.6           & 100.0          &  65.5           & 97.8           & 99.5          \\ 
$\textrm{CLIP}_{\textrm{L14}}$                     &  65.6          & 99.4             & 99.9           & 65.2              & 97.6            & 99.1          \\
VSE$\infty$              &  67.8          & 99.7          & 99.9          & 70.8          & 99.0          & 99.5          \\
\hline
$\textrm{VITR}_{\textrm{B}}$                  & 88.3 & 99.7 & \textbf{100.0} & \textbf{79.4} & 99.4 & 99.8 \\ 
$\textrm{VITR}_{\textrm{L}}$                  & \textbf{90.7} & \textbf{99.9} & 99.9 & 79.3 & \textbf{99.5} & \textbf{99.8} \\
\hline
\end{tabular}
% }
\end{table}

\begin{table}[ht]
\caption{Results of cross-modal information retrieval networks on the Flickr30K test set. Table shows average Recall@$K$ (\%) values.}
\label{Flickr30KResults}
\centering
% \resizebox{\linewidth}{!}{
\begin{tabular}{lcccccc}
\hline
\multirow{2}{*}{Network} & \multicolumn{3}{c}{Image-to-Text} & \multicolumn{3}{c}{Text-to-Image} \\
\cmidrule(lr){2-4} \cmidrule(lr){5-7}  
                         & Recall@1           & Recall@5           & Recall@10          & Recall@1           & Recall@5           & Recall@10          \\  \hline
VSE++ \cite{faghri2018vse++}                  & 52.9               & 80.5              & 87.2               & 39.6              & 70.1              & 79.5              \\
PFAN++ \cite{wang2020pfan++}                  & 70.1                & 91.8               & 96.1                & 52.7               & 79.9               & 87.0              \\
% CAAN \cite{zhang2020context}                   & 70.1          & 91.6          &  97.2          & 52.8          & 79.0           & 87.9          \\
% MMCA \cite{wei2020multi}                   & 74.2          & 92.8          & 96.4          & 54.8          & 81.4          & 87.8          \\
% MLSL \cite{li2021multi}                   & 72.2            & 92.4            & 98.2           & 56.8            & 83.3            & 91.3          \\
% RAAN \cite{wang2023rare}                  & 77.1            & 93.6         & 97.3          & 56.0           & 82.4 
         % & 89.1          \\
VSRN++ \cite{li2022image}                  & 79.2          & 94.6          & 97.5          & 60.6          & 85.6          & 91.4          \\   
Unicoder \cite{li2020unicoder}              & 86.2          & 96.3          & 99.0          & 71.5          & 90.9          & 94.9          \\
Uniter \cite{chen2020uniter}                   & 87.3          & 98.0          & 99.2          & 75.6          & 94.1          & 96.8          \\
ERNIE-ViL \cite{yu2021ernie}               & 88.7          & 98.0          & 99.2          & 76.7          & 93.6          & 96.4          \\
ViSTA-L \cite{cheng2022vista}             & 89.5         & 98.4          & 99.6          & 75.8          & 94.2          & 96.9          \\
$\textrm{CLIP}_{\textrm{RN101}}$                     & 88.3            & 98.2           & 99.4          & 72.9          & 92.6          & 96.2          \\
VILLA \cite{gan2020large}                   & 87.9          & 97.5          & 98.8          & 76.3          & 94.2          & 96.8          \\
VSE$\infty$ \cite{chen2021learning}            & 88.7          & 98.9          & 99.8          & 76.1          & 94.5          & 97.1          \\
$\textrm{CLIP}_{\textrm{B16}}$                     & 91.2           & 98.9           & 99.4          & 77.0          & 94.1          & 97.4          \\ 
$\textrm{COTS}^{\dagger }$ \cite{lu2022cots}                 & 91.7           & 99.0          & 99.9          & 78.3           & 94.9           & 97.2          \\
$\textrm{CLIP}_{\textrm{L14}}$                     & 92.6            & 99.2            & 99.6          & 77.8           & 95.2           & 97.7          \\ 
\hline
$\textrm{VITR}_{\textrm{B}}$                  & 93.7 & 99.1 & 99.8 & 80.8 & 95.7 & 97.9 \\ 
$\textrm{VITR}_{\textrm{L}}$                 & \textbf{94.7} & \textbf{99.7} & \textbf{99.9} & \textbf{82.5} & \textbf{96.7} & \textbf{98.3} \\
\hline
\end{tabular}
% }
\end{table}

\textbf{Results on Flickr30K.}
Extensive experiments were conducted to evaluate the performance of VITR on the widely-used benchmark dataset, Flickr30K. 
Table~\ref{Flickr30KResults} reveals the performance of the proposed VITR on the Flickr30K test set for cross-modal information retrieval, and the main findings are described as follows.
$\textrm{VITR}_{\textrm{L}}$'s Recall@1 values for image-to-text and text-to-image retrieval achieved 94.7\,\% and 82.5\,\% respectively. $\textrm{VITR}_{\textrm{L}}$'s Recall@1 values outperformed $\textrm{CLIP}_{\textrm{L14}}$'s Recall@1 values by 2.1\,\% and 4.7\,\% for image-to-text and text-to-image retrieval respectively.
Furthermore, $\textrm{VITR}_{\textrm{L}}$'s Recall@1 values also outperformed $\textrm{COTS}^{\dagger }$'s Recall@1 values by 3.0\,\% for image-to-text and 4.2\,\% for text-to-image retrieval. $\textrm{VITR}_{\textrm{B}}$ reached the Recall@1 of 93.7\,\% and 80.8\,\% for image-to-text and text-to-image retrieval tasks respectively. 
The results of Recall@1 of $\textrm{VITR}_{\textrm{B}}$ outperformed that of $\textrm{CLIP}_{\textrm{B16}}$ by 2.5\,\% and 3.8\,\% for image-to-text and text-to-image retrieval respectively.

\subsection{Results of VITR using the Turbo Module}

\textbf{Comparison of Retrieval Time between VITR with and without Turbo.}
Table~\ref{RetrievalTime} compares the retrieval times of $\textrm{VITR}_{\textrm{L}}$ (with and without the turbo module) to UNITER, when these are applied to the RefCOCOg test set. Here, $N$ is the number of selected candidates by turbo (see section \ref{turbo}), in the Table.
For retrieval of relevant descriptions from a pool of 9582 using a single query image, the average retrieval time of $\textrm{VITR}_{\textrm{L}}$ with turbo ($N=200$) is 0.3\,s, which is 13.7\,s faster than that without turbo and 10.5\,s faster than that of UNITER.
For retrieval of relevant images from a pool of 2600 using a single query description, the average retrieval time of $\textrm{VITR}_{\textrm{L}}$ with turbo ($N=200$) is 0.1\,s, which is 1.7\,s faster than that without turbo and 4.6\,s faster than that of UNITER.

\begin{table}[ht]
\caption{Comparison of the retrieval time of different models, including $\textrm{VITR}_{\textrm{L}}$ with and without turbo, and UNITER, using the RefCOCOg test set.}
\label{RetrievalTime}
\centering
% \resizebox{\linewidth}{!}{
\begin{tabular}{lcccc}
\hline
Task          & \begin{tabular}[c]{@{}c@{}}$\textrm{VITR}_{\textrm{L}}$\\ (turbo $N=200$)\end{tabular} & \begin{tabular}[c]{@{}c@{}}$\textrm{VITR}_{\textrm{L}}$\\ (turbo $N=500$)\end{tabular} & \begin{tabular}[c]{@{}c@{}}$\textrm{VITR}_{\textrm{L}}$\\ (without turbo)\end{tabular} & UNITER \\ \hline
Image-to-Text & 0.3\,s   & 0.8\,s   & 14.0\,s & 10.8\,s   \\
Text-to-Image & 0.1\,s   & 0.3\,s   & 1.8\,s & 4.7\,s   \\ \hline
\end{tabular}
% }
\end{table}

\begin{table}[ht]
\caption{Results of VITR with turbo for cross-modal information retrieval on the RefCOCOg test set. Table shows average Recall@$K$ (\%) values.}
\label{AblationStudiesturboModule}
\centering
% \resizebox{\linewidth}{!}{
% \begin{tabular}{l|ccc|ccc}
% \hline
% \multirow{2}{*}{turbo $N$} & \multicolumn{3}{c|}{Image-to-Text}            & \multicolumn{3}{c}{Text-to-Image}             \\ \cline{2-7} 
\begin{tabular}{lcccccc}
\hline
\multirow{2}{*}{Turbo $N$} & \multicolumn{3}{c}{Image-to-Text} & \multicolumn{3}{c}{Text-to-Image} \\
\cmidrule(lr){2-4} \cmidrule(lr){5-7} 
                         & Recall@1           & Recall@5           & Recall@10          & Recall@1           & Recall@5           & Recall@10          \\  \hline
100              & 45.2 & 71.1            & 80.3          & 29.5          & 55.1        & 66.7           \\

200              & 45.2 & 71.1            & 80.5          & 29.5          & 55.1        & 66.8           \\

500              & 45.2 & 71.1            & 80.5          & 29.5          & 55.1        & 66.8           \\ 

without turbo             & 45.2 & 71.1            & 80.5          & 29.5          & 55.1        & 66.8           \\
\hline
\end{tabular}
% }
\end{table}

\textbf{The Retrieval Performance of VITR with Turbo.} 
This section evaluates the impact of the proposed turbo module on the retrieval performance of VITR using the RefCOCOg test set. Table~\ref{AblationStudiesturboModule} shows that, for image-to-text and text-to-image retrieval, the retrieval performance of VITR using turbo with $N$ set to 200 and 500 is the same as that of VITR without turbo. When $N$ is set to 100, VITR with turbo underperformed VITR without turbo with a difference of 0.2\,\% for image-to-text retrieval and 0.1\,\% for text-to-image retrieval on Recall@10. 
The results in Table~\ref{RetrievalTime} and \ref{AblationStudiesturboModule} suggest that VITR with the proposed turbo ($N \geqslant 200$) achieved the same retrieval performance as VITR without turbo, but in a faster retrieval time.

\subsection{Ablation Studies on the Fusion}

% This section conducts ablation studies to evaluate the impact of fusing pre-trained knowledge and the results of relational reasoning in the proposed VITR network on relation-focused cross-modal information retrieval. 
This section undertakes a series of ablation studies to assess the influence of integrating pre-trained knowledge and the results of relational reasoning within the proposed VITR network. 
Experiments were carried out by creating variants of $\textrm{VITR}$ ($\textrm{VITR}_{\textrm{L}}$ model) and applying those to the RefCOCOg test set. The results of the experiments are shown in Table~\ref{AblationStudies}.

\begin{table}[ht]
\caption{Results of ablation studies on VITR's variant networks for cross-modal information retrieval on the RefCOCOg test set. Table shows average Recall@$K$ (\%) values.}
\label{AblationStudies}
\centering
\resizebox{\linewidth}{!}{
\begin{tabular}{lccccccc}
\hline
\multirow{2}{*}{Network}  & \multirow{2}{*}{Method}  & \multicolumn{3}{c}{Image-to-Text} & \multicolumn{3}{c}{Text-to-Image} \\
\cmidrule(lr){3-5} \cmidrule(lr){6-8} 

                        & & Recall@1           & Recall@5           & Recall@10          & Recall@1           & Recall@5           & Recall@10          \\  \hline

VSE$\infty$ & baseline             & 31.1          & 58.3            & 69.7          & 19.5          & 42.8         & 55.2           \\

$\textrm{CLIP}_{\textrm{L14}}$ & baseline                    & 42.4            & 65.5           & 75.1         & 25.2           & 48.9           & 60.4         \\ 

$\textrm{VITR-NoViT}$  & remove ViT            & 36.1          & 61.6           &  72.2          & 24.3         & 49.1          & 60.8          \\

$\textrm{VITR-NoRel}$ & remove RR                    & 43.1          & 66.7            & 76.9          & 25.3           & 49.3           & 60.4          \\
$\textrm{VITR}$     &  original         & \textbf{45.2} & \textbf{71.1} & \textbf{80.5} & \textbf{29.5} & \textbf{55.1} & \textbf{66.8} \\ \hline
\end{tabular}
}
\end{table}

The aim of the first experiment is to evaluate the performance of VITR when it does not utilise the ViT's image global representation, thereby assessing the impact of fusing the image global representation using the fusion module on the network. For this experiment, a new variant of $\textrm{VITR}$ was created, namely $\textrm{VITR-NoViT}$, that removes the ViT encoder. 
$\textrm{VITR-NoViT}$ outperformed VSE$\infty$ for image-to-text and text-to-image on Recall@1, with average improvements of 5.0\,\% and 4.8\,\% respectively. In addition, $\textrm{VITR-NoViT}$ underperformed $\textrm{VITR}$ by 9.1\,\% for image-to-text retrieval and 5.2\,\% for text-to-image retrieval on Recall@1.

The aim of the second experiment is to evaluate the performance of VITR when the relational reasoning module is removed, thereby assessing the impact of excluding the results of relational reasoning fusion on the network. 
For this experiment, a new variant of $\textrm{VITR}$ was created, namely $\textrm{VITR-NoRel}$ that does not include the relational reasoning module. The relational reasoning module was replaced by two GRUs, one for pooling the text and another for pooling the region features of images. The results of Recall@1 of $\textrm{VITR-NoRel}$ outperformed that of $\textrm{CLIP}_{\textrm{L14}}$ by 0.7\,\% and 0.1\,\% respectively, and the results suggest that the observed improvement is a result of ViT's pre-trained global knowledge being integrated into the network along with the results obtained from GRUs. Furthermore, it was observed that the performance of $\textrm{VITR-NoRel}$ was worse than that of $\textrm{VITR}$ by 2.1\,\% and 4.2\,\% for Recall@1 in image-to-text and text-to-image retrieval tasks, respectively. 

\section{Analysis}
\label{sec:visualisation}

\subsection{Visually Representing the Relational Reasoning Performance of VITR}
\begin{figure*}[htbp]
\centering
\includegraphics[width=1\linewidth]{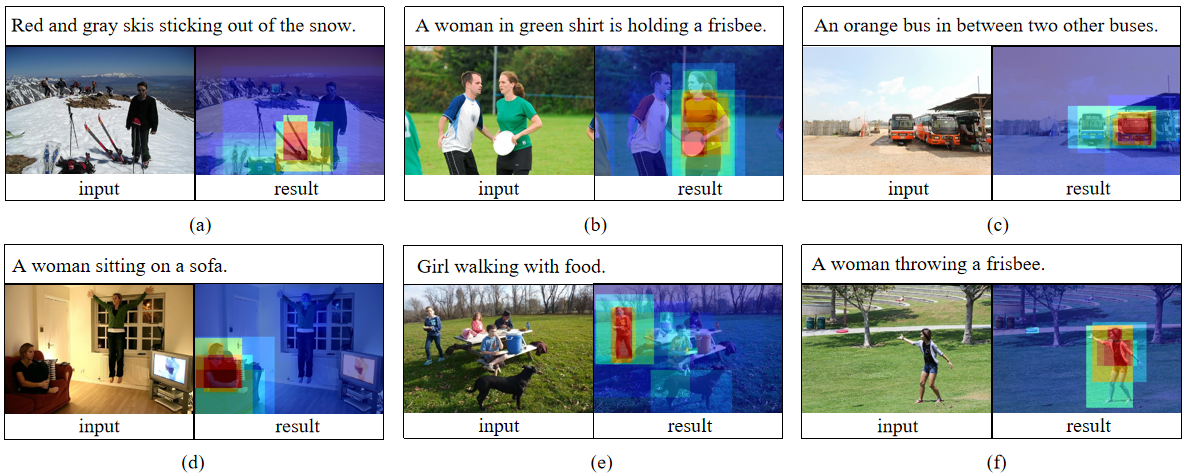}
\caption{
Visually representing the relational reasoning performance of VITR. In this figure, given a textual query and an input image, the visualisation is generated by highlighting the relevant image regions and darkening the irrelevant image regions.
}
\label{visualisation1}
\end{figure*}

Figure~\ref{visualisation1} presents an example visualisation of relational reasoning generated by the proposed VITR. In Figure~\ref{visualisation1}, the heat map highlights the image regions relevant to the textual query, and it is generated by the relational reasoning module as follows. 

Set $\{a_{i1}, \dotsc, a_{in}\}$, see Equation~\refeq{equationaij}, holds the weights for the $i$th image region, so let  $\bar{a}_i$ denote the average value the set. Let set $\{\bar{a}_1, \dotsc, \bar{a}_k\}$ holds the values of all image regions, and let its min-max normalisation result be $\{\bar{a}'_1, \dotsc, \bar{a}'_k\} \in [0,1]$. Therefore, $\bar{a}'_{i}$ is used as the heat degree for the $i$th image region.

The images from Figure~\ref{visualisation1}a--f show that the image regions only received focus by the relational reasoning module when they were mentioned in the query description. For example, in Figure~\ref{visualisation1}b, the image regions relevant to $\langle \text{`woman'}, \text{`holding'}, \text{`frisbee'}\rangle$ were the focus, while the other main region `man' in the image was ignored because it is irrelevant to the query description. The results of Figure~\ref{visualisation1} visually show the relational reasoning performance in VITR.

\begin{figure*}[htbp]
\centering
\includegraphics[width=1\linewidth]{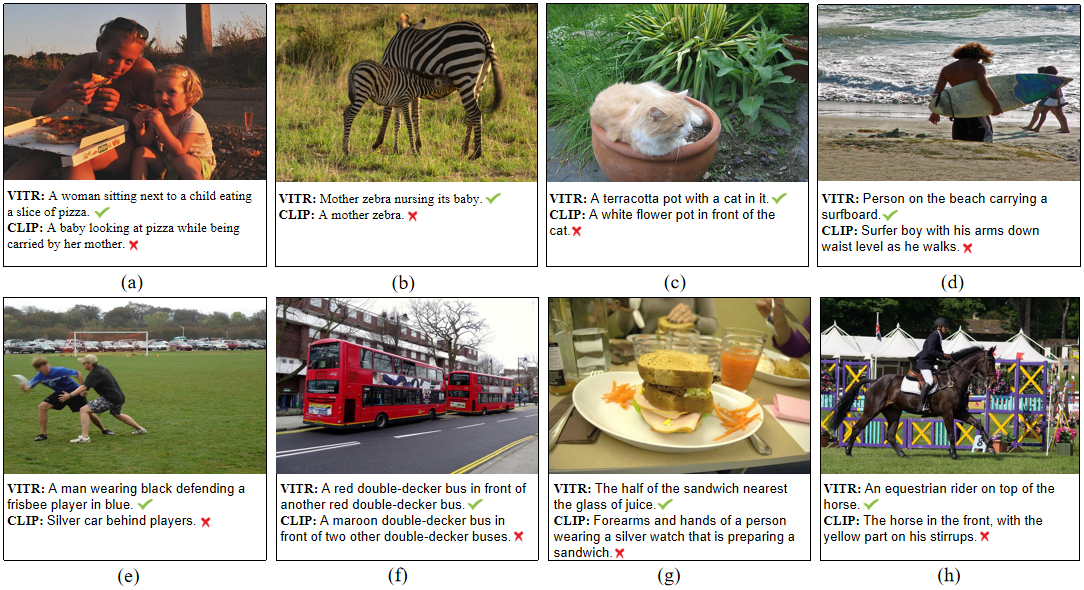}
\caption{
A comparison of the top one results for image-to-text retrieval using CLIP and VITR. CLIP's retrieved descriptions including the details do not match the query image, while VITR's retrieved descriptions concentrate on specific details of the image.
}
\label{visualisation2}
% \vspace{-35pt} 
\end{figure*}

\begin{figure*}[htbp]
\centering
\includegraphics[width=1\linewidth]{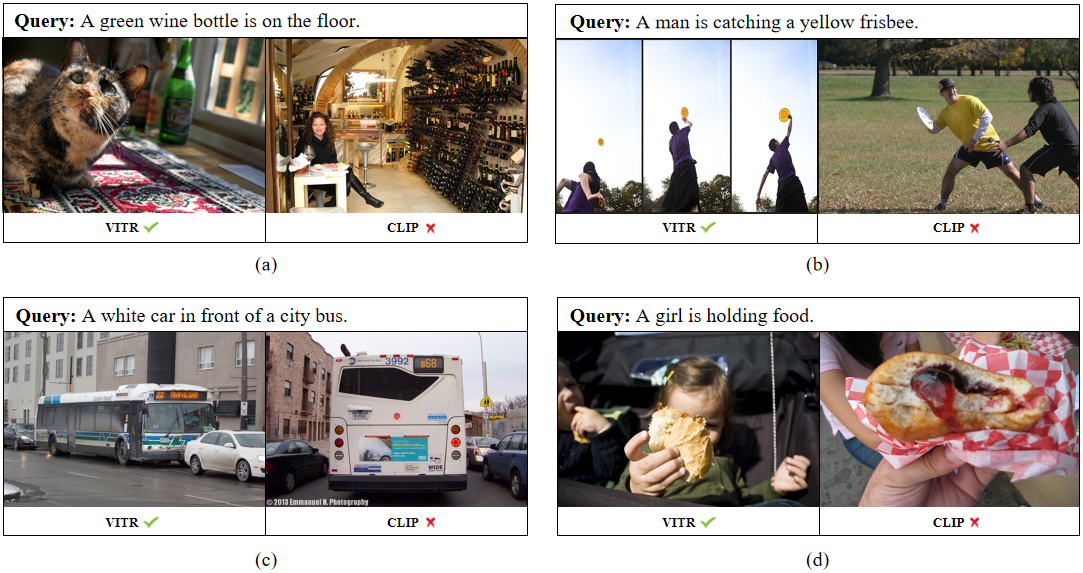}
\caption{
A comparison of the top one results for text-to-image retrieval using CLIP and VITR. None of CLIP's results align with the description, whereas VITR's results are more relevant to the query description.
}
\label{visualisation3}
\end{figure*}

\subsection{Comparison of Retrieval Results between CLIP and VITR}

\textbf{Image-to-Text Retrieval.} Figure~\ref{visualisation2} presents eight examples of the top one image-to-text retrieval results between CLIP and VITR. Typically, CLIP's results offer a description of the image with error or missing details of relations, while VITR's results concentrate on specific details. As seen in Figure~\ref{visualisation2}b, the result of CLIP describes the image as `A mother zebra' without mentioning relations, while the result of VITR describes it as a relation-focused sentence which is `A mother zebra nursing its baby'. Figure~\ref{visualisation2} highlights the limitations of CLIP in matching local image information, particularly relations, during image-to-text retrieval, and the improvement of VITR.

\textbf{Text-to-Image Retrieval.} Figure~\ref{visualisation3} presents four examples of the top one results of text-to-image retrieval between CLIP and VITR. As shown in Figure~\ref{visualisation3}c, the query aims to find an image of a white car in front of a bus, but the result from CLIP includes errors in the relations between the car and the bus, making the retrieved image less relevant to the query. On the other hand, VITR produces more accurate results that are better aligned with the intent of the query. Figure~\ref{visualisation3} highlights the limitations of CLIP in matching relation information between images and descriptions during text-to-image retrieval, and the improvement of VITR.

\section{Conclusion}
\label{sec:conclusion}
This paper presents an innovative network that combines the local representations of an image with its global representation derived from the ViT model. The proposed network, VITR, is specifically designed for enhancing cross-modal information retrieval tasks. 
VITR includes a relational reasoning module that extends the capabilities of ViT by modeling the relations of regions in images for relation-focused cross-modal information retrieval; a fusion module that fuses the image global information from the ViT and the relation reasoned information of relational reasoning. 
Empirical evaluations revealed that the proposed VITR network outperformed CLIP and other VSE networks for both relation-focused and traditional cross-modal information retrieval tasks. 
When assessed through the average Recall@1 evaluation metric for retrieval performance, VITR exhibited superior results compared to CLIP. On the RefCOCOg dataset, VITR outperformed CLIP by 2.8\% for image-to-text retrieval and 4.3\% for text-to-image retrieval. On the CLEVR dataset, VITR achieved a substantial improvement of 25.1\% for image-to-text retrieval and 14.1\% for text-to-image retrieval. Similarly, on the Flickr30K dataset, VITR showed improvements of 2.1\% for image-to-text retrieval and 4.7\% for text-to-image retrieval when compared to CLIP.
While the proposed VITR network is effective in image-to-text and text-to-image retrieval tasks, its limitation is that it does not consider other similar tasks, such as video-to-text retrieval, which are essential in many applications. To overcome this limitation, future research could focus on developing cross-modal neural networks capable of handling multiple tasks, thereby providing solutions for a broader range of applications.

%Bibliography
\bibliographystyle{unsrt}  
\bibliography{egbib}  

\end{document}